%% file: main.tex
\documentclass[10pt,twocolumn,letterpaper]{article}
\usepackage[pagenumbers]{cvpr} 
\usepackage[numbers]{natbib}
\usepackage{graphicx}
\usepackage{amsmath}
\usepackage{amssymb}
\usepackage{booktabs}
\usepackage[utf8]{inputenc} 
\usepackage[T1]{fontenc}    
\usepackage{hyperref}       
\usepackage{url}            
\usepackage{booktabs}       
\usepackage{amsfonts}       
\usepackage{nicefrac}       
\usepackage{microtype}      
\usepackage{xcolor}         
\usepackage{graphicx}
\usepackage{algorithm}
\usepackage{algorithmic}
\usepackage{tabularray}
\usepackage{multirow}
\usepackage{amsmath}
\usepackage{ulem}
\usepackage{float}
\usepackage{adjustbox}
\usepackage{multicol}
\UseTblrLibrary{booktabs}
\definecolor{cBlue}{rgb}{0.6,0.8,0.9}
\definecolor{cGray}{rgb}{0.85,0.85,0.85}

\usepackage[capitalize]{cleveref}
\crefname{section}{Sec.}{Secs.}
\Crefname{section}{Section}{Sections}
\Crefname{table}{Table}{Tables}
\crefname{table}{Tab.}{Tabs.}

\def\cvprPaperID{*****} 
\def\confName{CVPR}
\def\confYear{2022}

\begin{document}

\title{VEGA\raisebox{0.7ex}{\includegraphics[scale=0.05]{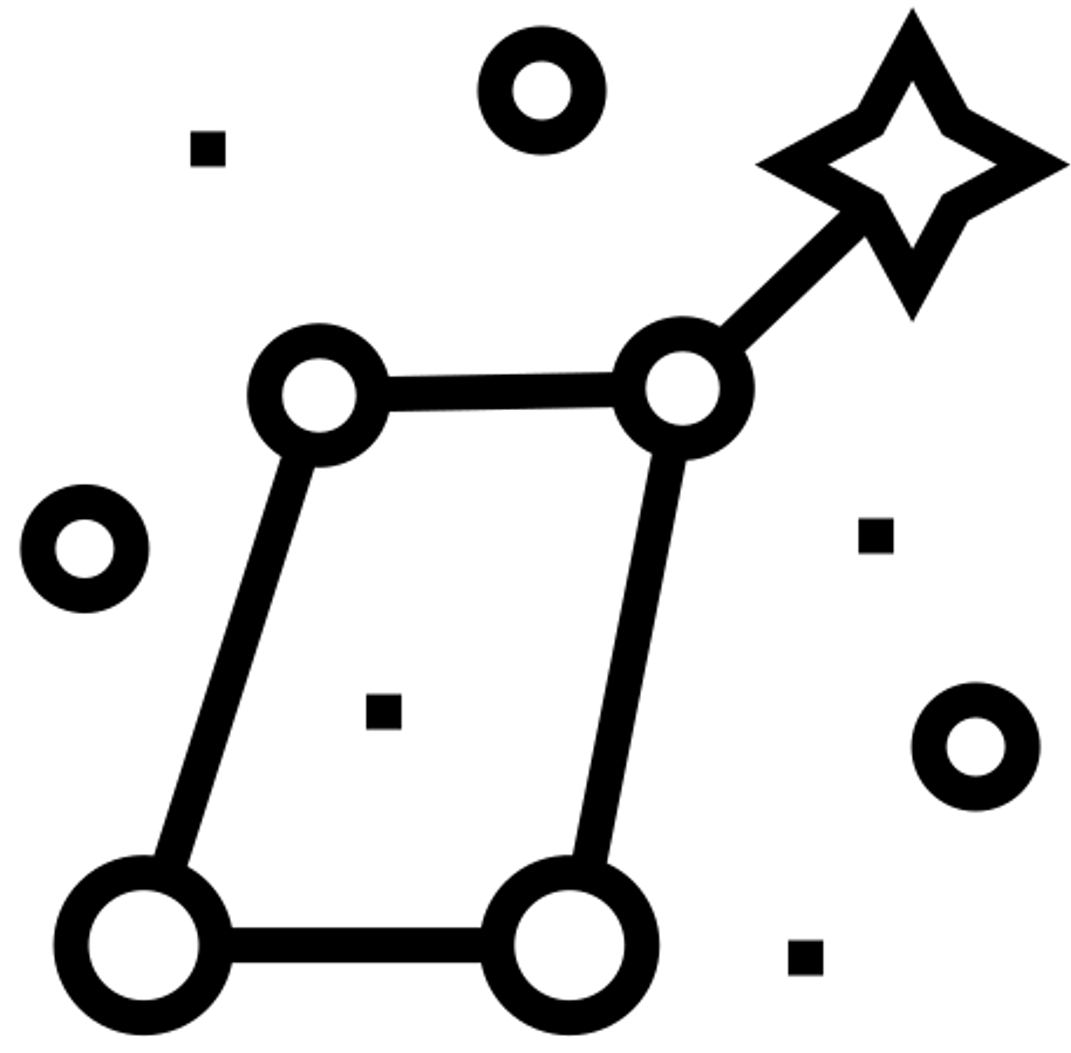}}: Learning Interleaved Image-Text Comprehension in Vision-Language Large Models}

\author{
    Chenyu Zhou$^{1,\ast}$, Mengdan Zhang$^\ast$, Peixian Chen$^{\spadesuit,\ast}$, Chaoyou Fu, Yunhang Shen \\ 
    Xiawu Zheng$^{1,\dagger}$,  Xing Sun, Rongrong Ji$^{1}$  \\
    \\
    {\small$^{1}$Key Laboratory of Multimedia Trusted Perception and Efficient Computing,} \\
    {\small Ministry of Education of China, Xiamen University} \\
    {\tt\small zhoucy977@stu.xmu.edu.cn, 
\{pxchen13, zhangmengdanrz\}@gmail.com}
}

\maketitle

\renewcommand{\thefootnote}{\fnsymbol{footnote}}
\renewcommand{\thefootnote}{\fnsymbol{footnote}}
\footnotetext[1]{Equal Contribution}
\footnotetext[2]{Corresponding Author}
\renewcommand{\thefootnote}{$\spadesuit$}
\footnotetext[3]{Project Leader}

\input{sec/0_abstract}    
\input{sec/1_intro}

\input{sec/2_related}
\input{sec/3_methodv2}

\input{sec/4_experiments}

\input{sec/5_conclusion_limitation}

{
    \small
    \bibliographystyle{plainnat}
    \bibliography{main}
}
\input{sec/6_appendix}

\end{document}

%% file: sec/0_abstract.tex
\newcommand{\SIAQA}{SIAQA}
\begin{abstract}
\label{abstract}

The swift progress of Multi-modal Large Models (MLLMs) has showcased their impressive ability to tackle tasks blending vision and language.
Yet, most current models and benchmarks cater to scenarios with a narrow scope of visual and textual contexts.
These models often fall short when faced with complex comprehension tasks, which involve navigating through a plethora of irrelevant and potentially misleading information in both text and image forms.
To bridge this gap, we introduce a new, more demanding task known as Interleaved Image-Text Comprehension (IITC).
This task challenges models to discern and disregard superfluous elements in both images and text to accurately answer questions and to follow intricate instructions to pinpoint the relevant image.
In support of this task, we further craft a new VEGA dataset, tailored for the IITC task on scientific content, and devised a subtask, Image-Text Association (ITA), to refine image-text correlation skills.
Our evaluation of four leading closed-source models, as well as various open-source models using VEGA, underscores the rigorous nature of IITC.
Even the most advanced models, such as Gemini-1.5-pro and GPT4V, only achieved modest success.
By employing a multi-task, multi-scale post-training strategy, we have set a robust baseline for MLLMs on the IITC task, attaining an $85.8\%$ accuracy rate in image association and a $0.508$ Rouge score. These results validate the effectiveness of our dataset in improving MLLMs capabilities for nuanced image-text comprehension. Project page: \url{https://zhourax.github.io/VEGA/}

\end{abstract}

%% file: sec/1_intro.tex
\section{Introduction}
\label{sec:intro}

The swift advancement of Multi-modal Large Language Models (MLLMs)~\cite{openai2024gpt4,geminiteam2024gemini,chen2023internvl,liu2023llava,li2023blip2} has recently showcased their remarkable aptitude for tackling vision-language tasks.
These models have significantly improved in areas such as logical reasoning~\cite{yue2023mmmu,han2023infimmeval}, high-resolution image comprehension~\cite{internlmxcomposer2_4khd}, In-Context Learning (ICL)~\cite{zeng2024mllms,baldassini2024makes}, and Chain of Thought (COT) processing~\cite{ge2023chain}, leading to notable achievements.
Despite these advancements, current applications typically engage MLLMs in the basic multi-modal comprehension task, which involves posing questions tied to a narrow scope of visual content.
This approach falls short when compared to human daily comprehension.

\begin{figure*}
  \centering
  \includegraphics[width=1.0\linewidth]{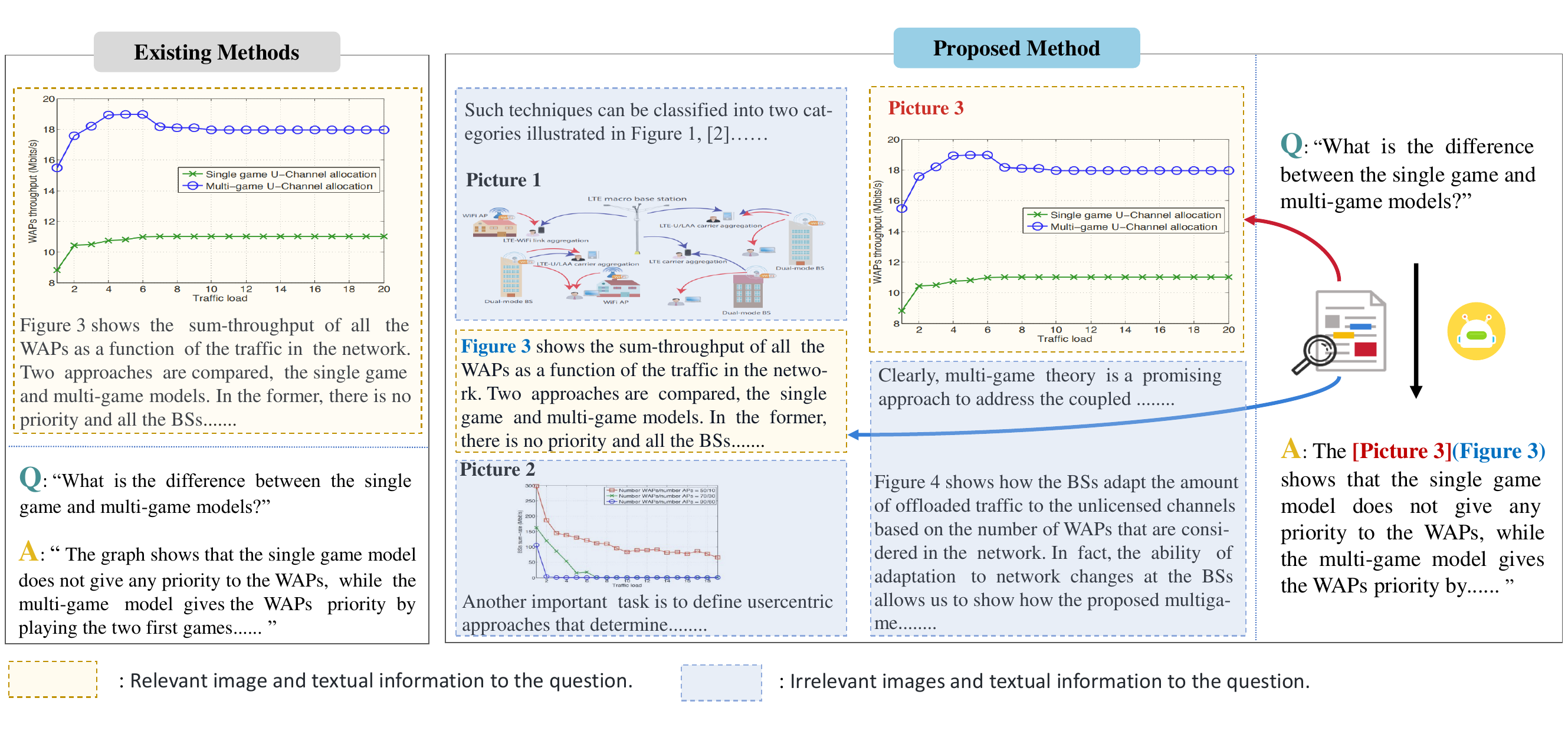}
  \caption{Comparison between existing VQA tasks and our IITC task. \colorbox{cGray}{Left}: The input for existing VQA tasks only incorporates a limited amount of image and text data, which is highly relevant to the question. \colorbox{cBlue}{Right}: The input for the IITC task contains longer images and text information, which includes redundant and misleading data. The model needs to specify the reference image when providing an answer. }\label{fig:intro1}
\end{figure*}

Human comprehension, particularly in document analysis, presents unique challenges not found in basic comprehension tasks such as Visual Question Answering (VQA) tasks~\cite{antol2015vqa}. As shown in Fig.~\ref{fig:intro1}, these challenges include: 1) \textbf{Interleaved Text-Image Distraction}: MLLMs are forced to find the correct context from irrelevant text-image mixtures.
2) \textbf{Long-Form Content}: The inputs consist of intertwined sequences of images and text, demanding comprehension over long multimodal context.
3) \textbf{Image-Referenced Answers}: The model is tasked with identifying and linking the relevant image and text based on specific instructions before crafting an appropriate response. Recent prominent open-source MLLMs, such as LLaVA 1.5~\cite{liu2023llava}, CogVLM~\cite{wang2023cogvlm}, and PaperOwl~\cite{hu2023paperowl}, lack support for multiple image inputs and struggle with such intricate comprehension task. Multi-image, multi-turn models~\cite{chen2023internvl} and video-based MLLMs~\cite{li2024videochat} face challenges in concurrently processing extended visual and textual contexts effectively. Additionally, benchmarks for evaluating such an intricate task are limited, obscuring our understanding of MLLMs' practical comprehension abilities.

To bridge this gap, we introduce a novel multi-modal task, Interleaved Image-Text Comprehension (IITC), which involves locating the relevant text and image within a complex context given a question, providing an accurate answer, and outputting the corresponding image index. Meanwhile, we present the first benchmark MLLM to handle the above challenges within the IITC task.

Specifically, we observe that IITC is a complex task where simply increasing the number of Interleaved Image-Text QA samples is insufficient for accurate comprehension.
Therefore, we propose utilizing a multi-task learning strategy to enhance the model's understanding capabilities: employing the Image-Text Association (ITA) auxiliary task to strengthen the model's ability to accurately locate image-text paragraph correspondences.
In addition, we introduce a multi-scale training strategy, constructing the training data with progressively increasing context lengths and image numbers to enhance the model's robustness against gradually increasing redundant image-text content.

Ultimately, we have developed a novel dataset, designated as VEGA. It is comprised of two subsets, one tailored for the IITC task and another for the ITA task. The longest interleaved image-text content in VEGA reaches up to 8 images and 8k tokens. We design the instruction of the IITC task to be a question about only one of the images, requiring the model to specify the image it refers to in its answer.
We assess the model's interleaved image-text reading comprehension ability by both the correct rate of associated images, and the text quality of the answer by ROUGE\cite{lin2004rouge} and BLEU~\cite{papineni-etal-2002-bleu}. We have evaluated several state-of-the-art MLLMs on our dataset, validating the challenge of our tasks. Furthermore, we have fine-tuned the Qwen-VL-Chat model~\cite{Qwen-VL} on the VEGA dataset to set a robust baseline for the IITC task.

In conclusion, this study makes several noteworthy contributions to the field:
\begin{itemize}
    \item \textbf{Identifying a Novel Task.} We introduce a novel Interleaved Image-Text Comprehension (IITC) task. It evaluates and enhances MLLMs' capabilities of following complex instructions and extracting key cues in long and interleaved image-text scenarios.

    \item \textbf{Introducing the VEGA Dataset.} We develop a new VEGA Dataset for the IITC task that enables a comprehensive understanding of scientific literature, whose multi-modal context reaches up to 8,000 tokens in length and contains up to 8 images.

    \item \textbf{Benchmarking MLLMs in IITC.} We utilize the VEGA dataset to appraise the IITC capabilities of current state-of-the-art models, including GPT4V, Gemini-1.5-pro, and Qwen-VL-Chat, experimentally validating the challenge of IITC.

    \item \textbf{Enhancing MLLMs’ IITC Capabilities.} We fine-tune the Qwen-VL-Chat model on the VEGA dataset with a multi-scale, multi-task training strategy and our experimental findings demonstrate that it achieves an image association accuracy rate of 85.8\%. This represents a significant improvement and establishes a strong baseline for the IITC task.
\end{itemize}

%% file: sec/2_related.tex
\section{Related Work}

\subsection{MLLMs}
Recently, there has been a notable increase in the deployment of MLLMs aimed at tackling more complex tasks that involve multiple forms of media. These advanced models are equipped to understand and interpret not just text, but also visual content like images and videos~\cite{ge2023making}. For instance, BLIP-2 \cite{li2023blip} features the Q-Former, a key component that establishes a link between a static LLM and visual data, showing impressive results in VQA tasks. InstructBLIP \cite{dai2024instructblip} is specifically fine-tuned using a variety of instruction-based datasets, which improves its understanding of visual scenes and dialogues. Multi-modal-CoT \cite{zhang2023multimodal} brings the concept of chain-of-thought \cite{wei2022chain} into the multi-modal domain, demonstrating strong performance on the ScienceQA benchmark \cite{lu2022learn}.
LLaVA \cite{liu2024visual} uses a linear approach and fine-tunes the entire LLM to enhance its effectiveness. The LLaVA-NeXT \cite{liu2024llava}, compared to LLaVA-1.5, quadruples the pixel resolution of input images, improves visual guidance for data blending, and offers enhanced visual reasoning and OCR capabilities. 
The LLaMA-Adapter \cite{zhang2023llama} introduces a lightweight adapter module, enhancing the model's capability to process visual information. NExT-GPT \cite{wu2023next} connects LLM with multimodal adapters and various diffusion decoders, enabling NExT-GPT to perceive inputs and generate outputs in any combination of text, images, video, and audio.

In this work, our focus is on the IITC task, so we concentrate only on models that support multi-image input, such as Qwen-VL \cite{Qwen-VL} and InternVL \cite{chen2023internvl}. We use Qwen-VL-Chat as the base model for training to establish a baseline for the IITC task.

\begin{figure*}[ht]
  \centering
  \includegraphics[width=1.0\linewidth]{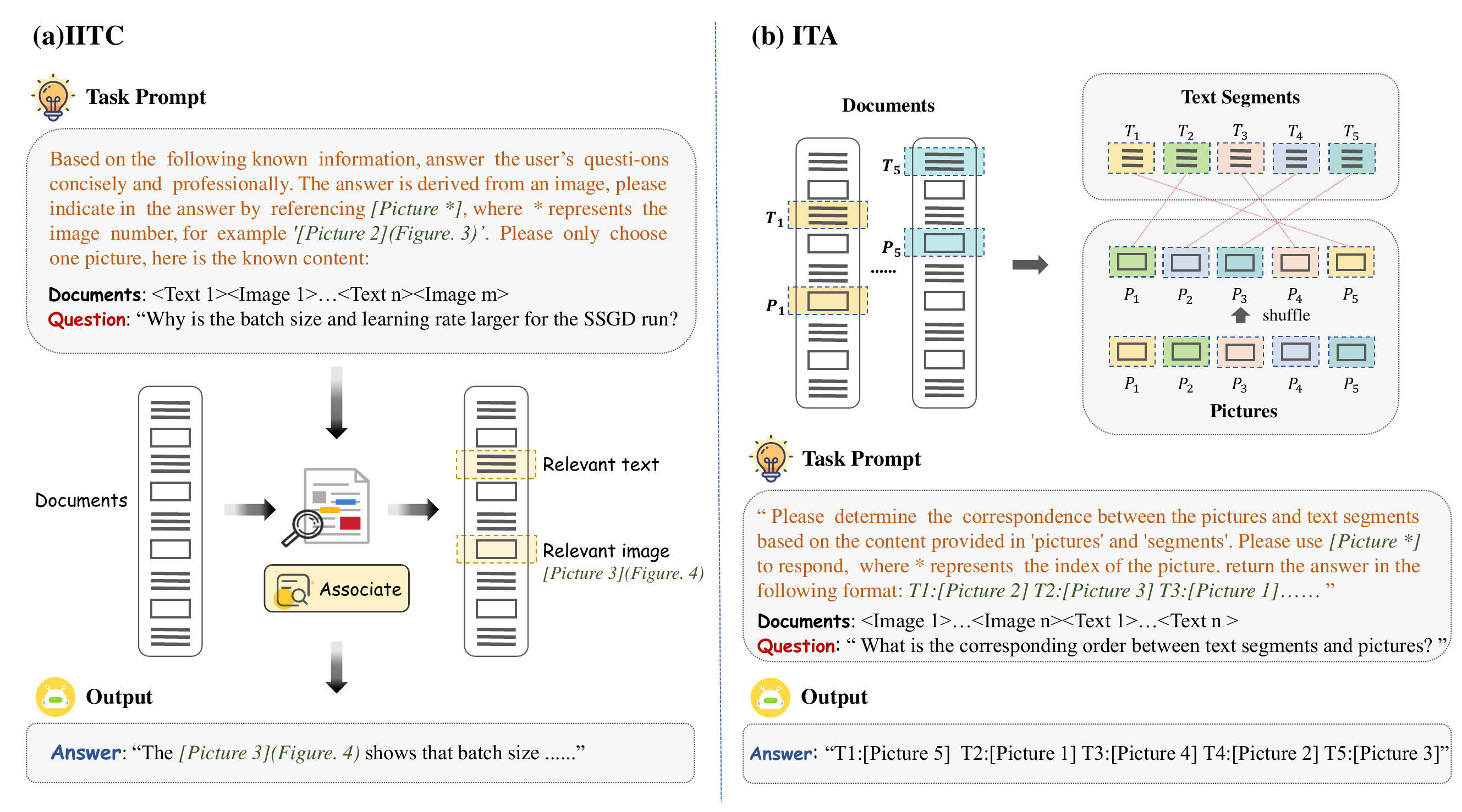}
  \caption{The task definition of IITC and ITA tasks. (a) The IITC task takes long interleaved image-text content as input and requires the model to specify the image it refers to in its response. (b) The ITA task takes shuffled images and text segments from different articles as input and requires the model to output the relationship between the text and the images. <Text *> and <Image *> represent a text segment and an image, respectively. They are both tokenized and fed into the model along with the task prompt and the question. }\label{fig:task_definition}
\end{figure*}

\subsection{MLLM Benchmarks}

The evolution of MLLMs, driven by multi-modal pretraining and instruction tuning, has outpaced traditional benchmarks like visual question answering and image captioning. New benchmarks have emerged to assess aspects such as OCR capabilities, adversarial robustness, and hallucination susceptibility. POPE \cite{li2023evaluating}, HaELM HaELM \cite{wang2023evaluation}, LAMM \cite{yin2023lamm}, LVLM-eHub \cite{xu2023lvlmehub}, and MM-Vet \cite{yu2023mmvet} provide insights into these areas and give a holistic view of MLLM's performance.

Vision-language benchmarks like Winoground \cite{thrush2022winoground}, RAVEN \cite{zhang2019raven} , OK-VQA \cite{marino2019ok}, and VCR \cite{zellers2019recognition} aim to measure MLLMs' nuanced reasoning abilities. TextVQA \cite{wright2010sparse}, FigureQA \cite{kahou2017figureqa}, ScienceQA \cite{lu2022learn}, and MathVista \cite{lu2023mathvista} contribute to understanding MLLMs' domain-specific reasoning. Comprehensive benchmarks like MME \cite{fu2024mme}, MMBench \cite{fu2024mme}, and SEED-Bench cover a variety of reasoning skills. However, the prevalent use of multiple-choice formats can lead models to rely on hints and overlook the reasoning process, focusing only on the final answer's correctness.

%% file: sec/3_methodv2.tex
\section{Proposed Method}

In this section, we introduce our VEGA dataset by first elaborating the task definition (\ref{sec: task_defi}), followed by outlining the process of construction (\ref{sec: data_collection}) and statistics (\ref{sec: data_statistics}). Finally, we present the training details of our baseline MLLM model (\ref{sec: baseline_model}).

\label{sec:method}
\subsection{Task Definition}
\label{sec: task_defi}

\textbf{Interleaved Image-Text Comprehension (IITC).} Our VEGA dataset is meticulously designed to address the challenging IITC task, with the goal of advancing the real-world comprehension capabilities of MLLMs. As demonstrated in Fig.~\ref{fig:intro1}, an MLLM is presented with the complex task of discerning relevant text and images in response to a given question.
The model is expected to not only deliver an accurate answer but also to identify the specific image index related to the response. This requirement extends beyond the traditional boundaries of VQA tasks~\cite{antol2015vqa}.

To facilitate this process, we have developed a specialized prompt that directs  MLLM to carry out the desired instruction. The complete input for MLLM is organized as shown in Fig. \ref{fig:task_definition} (a). We utilize the notation ``[Picture n]'' within the text to establish a clear reference to the n-th image, whether it appears in various sections of an article or across different examples. This approach enables the model to learn a consistent and standardized format for image referencing.

In the IITC task, the crux of the challenge lies in the model's proficiency in accurately linking the appropriate images with the corresponding textual information as dictated by the instructions. This precise pairing is crucial, for it is the foundation upon which a coherent answer is constructed. To quantitatively gauge the model's aptitude for this association, we have crafted each question pertaining to a specific input image, necessitating that the model explicitly reference this image as part of its response.
The model's prowess in image-text association is quantified by its success rate in identifying the correct image. Meanwhile, the textual response's quality is evaluated using established linguistic metrics, namely the ROUGE~\cite{lin2004rouge} and BLEU~\cite{papineni-etal-2002-bleu} scores. Together, these measures provide a comprehensive assessment of the model's performance on the IITC task.

\textbf{Image-Text Association (ITA)}. It is the auxiliary task designed to sharpen MLLM's precision in aligning text paragraphs with their relevant images. As illustrated in Fig. \ref{fig:task_definition}(b), ITA challenges the model with a shuffled array of inputs: $\left(P_1, \ldots, P_n, T_{1}, \ldots, T_n\right)$. Each image, labeled as $P_i$ from $i=1$ to $n$, is linked to a unique article, while each text snippet $T_{i}$  provides context for its image $P_{i}$. The textual inputs are intentionally scrambled at the outset. ITA's goal is for the model to accurately match and declare the correct pairings of each text segment $T_{i}$   with its image $P_{i}$, as demonstrated in Fig. \ref{fig:task_definition}(b).
Compared to the IITC task, ITA places greater emphasis on the model's ability to associate images with text. It is anticipated that training with ITA will bolster this skill within the model, leading to improved performance on the IITC task.
For the ITA task evaluation, with a set of  $n$ images, we count the number of correctly associated image-text pairs $m$, and use the ratio $m/n$ as the performance metric for the ITA task.

\begin{figure*}[t]
  \centering
  \includegraphics[width=1.0\linewidth]{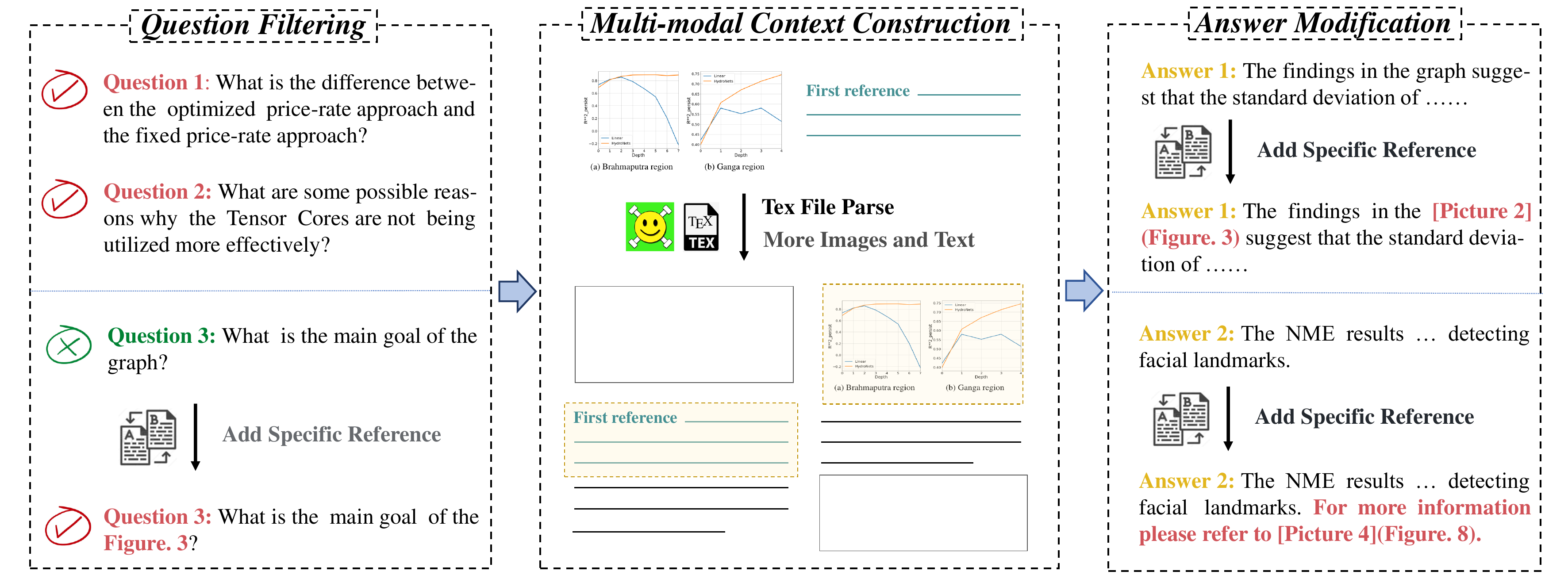}
  \caption{The construction process of the VEGA dataset. We use the SciGraphQA dataset as a foundation, to which we add more images and text to its context, modify questions and answers that lack clear image references, and incorporates references to meet the requirements of the IITC task. }\label{fig:IITC construction}
\end{figure*}

\subsection{Data Collection}
\label{sec: data_collection}

We develop the VEGA dataset upon the foundation of SciGraphQA~\cite{li2023scigraphqa}.
SciGraphQA is a multi-turn question-answering dataset for scientific graphs, containing 295k high-quality, multi-turn data entries.
It collected part of the papers published on arXiv from 2010 to 2020, and extracted the pictures (denoted by $\{P_i\}^N_{i=1}$, $N$ is the total number of pictures), the captions ($\{C_i\}^N_{i=1}$) of the pictures, and the first paragraphs ($\{M_i\}^N_{i=1}$) mentioning the pictures from the papers.

As outlined in Fig.~\ref{fig:IITC construction}, our enhancement involves a meticulous process of questions filtering that pertain to the visual-textual context, constructing long visual and textual context, and formulating answers that directly reference images.
The VEGA dataset is structured into two subsets, each specifically curated to train models on the IITC and ITA tasks, respectively.

\subsubsection{Subset for IITC} 
\label{subsec:Subset for IITC}

\textbf{Question Filtering}.
In the original SciGraphQA dataset, some questions about images are quite general, such as ``What are the implications of the results shown in the graph?'' and ``What does the x-axis and y-axis of the graph represent?'' For more examples, please refer to the supplementary materials. However, when these questions are placed within the context of multi-image long articles, they encounter issues with ambiguous image references. Furthermore, the answers are heavily dependent on the image content, rather than the multi-modal context information, which is not aligned with the objectives of the IITC task. To address this, as illustrated in Fig.~\ref{fig:IITC construction}, we modify the training set by reducing the proportion of such data and incorporating explicit image references in the questions, thereby enhancing the model's image-reading capabilities. For the test set, we filter out these questions and focus on evaluating the model's ability to comprehend complex interactions between text and images based on specific queries. Through manual screening, we curate over 2,000 questions and selecte 700 high-quality questions to form the final IITC test set.

\textbf{Context Construction}.
We investigate two approaches for crafting extended multi-modal contexts. The first approach involves randomly merging images and their introductory paragraphs from various papers in the original SciGraphQA dataset into a long context designated as $<P^k_iM^k_i><P^l_jM^l_j>...<P^m_nM^m_n>$ where $k$, $l$ and $m$ denote different papers, and then posing questions on each image and its textual surroundings. 
The second expands the text-image sequence from the original paper based on the paragraph $\{M_i\}^N_{i=1}$. The expanded sequence is denoted as $\{E_i\}^N_{i=1}$, which can maintain the structure of the paper and capture the inherent multi-modal context.

For the IITC task's test set, we utilize the latter approach. During the training set construction, we assess models trained with both approaches, noting superior performance with the second, as mentioned in Sec. \ref{subsec:Ablation Study}.
Beyond the advantage of consistency in data construction, we identify two key reasons for this outcome. First, the disparity in image-text content across different papers is significant, and the first approach does not effectively support the model's ability to discern between relevant and distracting multi-modal information, thereby hindering its capability to accurately match questions with the correct image-text responses. Second, the brevity of the first mention paragraph  $M_i$ often results in an insufficient textual context,  impeding question comprehension and response.
Consequently, we select the second one for assembling the multi-modal long context for IITC.

\begin{figure*}
  \centering
  \includegraphics[width=1.0\linewidth]{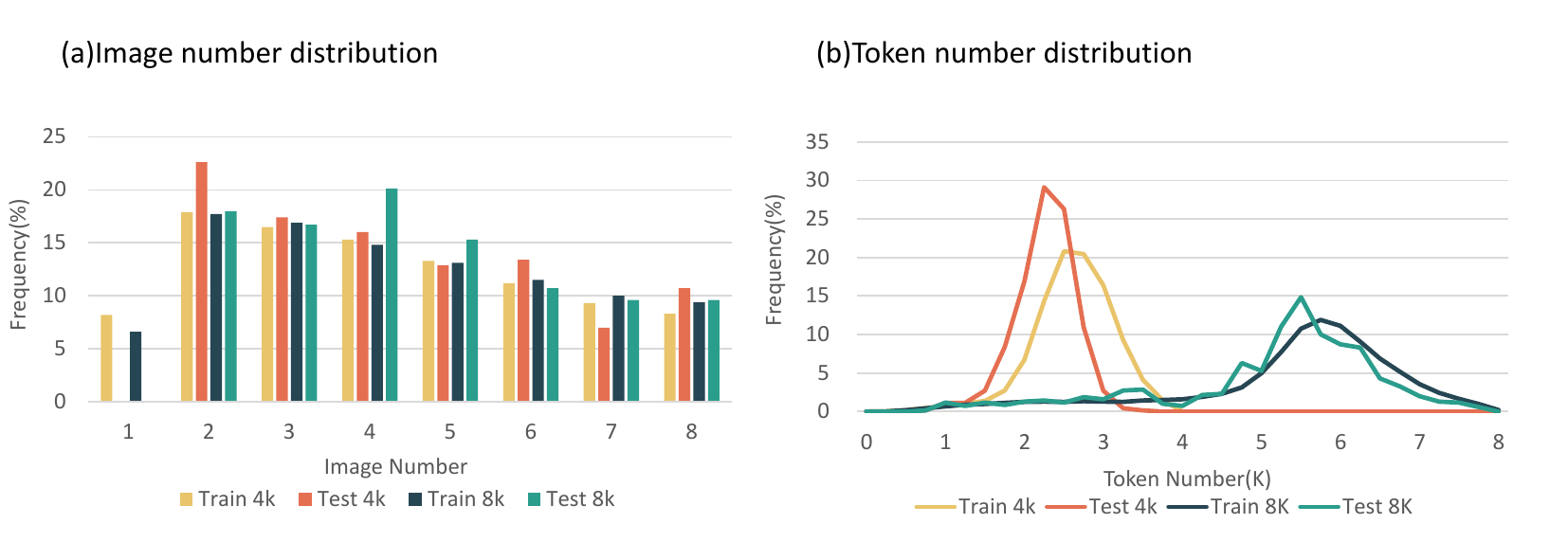}
  \caption{The distribution of the number of images and tokens in the IITC subset of the VEGA dataset. The number of tokens for each image is 256. }\label{fig:tokennum}
\end{figure*}

We elaborate on the second approach of context construction.
Illustrated in Fig.~\ref{fig:IITC construction}, 
we retrieve the paper's Tex files from Arxiv based on the `id' field from the original SciGraphQA dataset. We then match the 'caption' field with TeX commands `\textbackslash caption\{.\}', `\textbackslash label\{.\}', and `\textbackslash ref\{.\}' to pinpoint the paper's paragraph that first mentions the picture. Next, we expand this paragraph's context by sequentially incorporating lines from the adjacent text. We intersperse 2 to 8 random pictures from the document into this text until achieving a specific context length.
To evaluate the model's efficacy across varying token lengths, we craft two dataset versions capped at 4k and 8k tokens, ensuring a balanced token count distribution as depicted in Fig. \ref{fig:tokennum}.
To accurately represent the text-image spatial relationship, we maintain the sequence and placement of the inserted images as they appear in the TeX files. Lastly, we substitute the '\textbackslash ref\{.\}' markers related to the embedded images with the notation "[Picture n]", clarifying the link between these references and their corresponding images.

\textbf{Answer Modification}.
IITC requires the model to specify the image it refers to in its response. To meet this requirement, we modify the answers in the original dataset. We replace the subjects in the original answers, such as `graph' and `figure', with the notation [Picture $i$](Figure $j$), where $i$ denotes the sequence number of the input image, and $j$ signifies the image's index within the original document. Alternatively, we add `For more information, please refer to [Picture $i$](Figure $j$).'

\subsubsection{Subset for ITA}
We craft the Image-Text Association (ITA) task to bolster the model's proficiency in correlating text with images and to fortify its accuracy in image indexing responses. 
For each image, we generate a textual context $T_i = E_i$ akin to the IITC task, extending the text around the first mention paragraph $M_i$, capped at 2,000 characters. Note that in the ITA task, the expanded context paragraphs $E_i$ does not include images outside of $P_i$.
We streamline the ITA task by pairing images with text from disparate articles, randomizing them, and prompting the model to reorder these pairs. 
Despite the task's complexity, leading-edge methods like Qwen-MAX~\cite{QwenVLMax2024} and Gemini-1.0-pro~\cite{geminiteam2024gemini} exhibit only moderate success, particularly as the volume of pairs escalates, as evidenced in Table \ref{tab:main_result}. At the same time, we find that direct training with expanded image-text pairs is insufficient for the model to master text-image associations. To mitigate the learning challenge, we implement a multi-scale training strategy in the ITA task. We design three textual scales for the training set, corresponding to the image caption $C_i$, the first mention paragraph $M_i$, and the expanded context paragraphs $E_i$.  We also design two image quantity scales, with sets of three and five images. Experiments have shown that this multi-scale training set construction method can make the model more adaptable to the ITA task. In testing, we evaluate the model using solely the extended pairs, independent of other scaled data.

\begin{figure}[h]
    \centering
    \begin{minipage}{0.45\textwidth}
        \centering
        \begin{table}[H]
            \small
            \begin{tblr}{
              cell{2}{1} = {r=2}{},
              cell{4}{1} = {r=6}{},
            }
            \toprule
            \textbf{Subset} & \textbf{Scale}                     & \textbf{Train}    & \textbf{Test} & \textbf{Total} \\
            \midrule 
            IITC & 4k Tokens    & 208103  & 672 & 208775 \\
                 & 8k Tokens    & 196947 & 658 & 197605 \\
            \midrule
            ITA  & 3 Pic C.         & 23991      &-   &23991   \\
                 & 3 Pic M. & 23993       &-     &23993 \\
                 & 3 Pic E.  & 47983   & 500 &48483 \\
                 & 5 Pic C.         & 23986       &-  &23986    \\
                 & 5 Pic M. & 23985       &-     &23985 \\
                 & 5 Pic E.  & 45226  & 486  & 45712 \\
            \bottomrule 
            \end{tblr}
            \caption{The composition of the VEGA Dataset.}\label{table:composition_vega}
        \end{table}
    \end{minipage}\hfill
    \begin{minipage}{0.45\textwidth}
        \centering
        \includegraphics[width=1.0\linewidth]{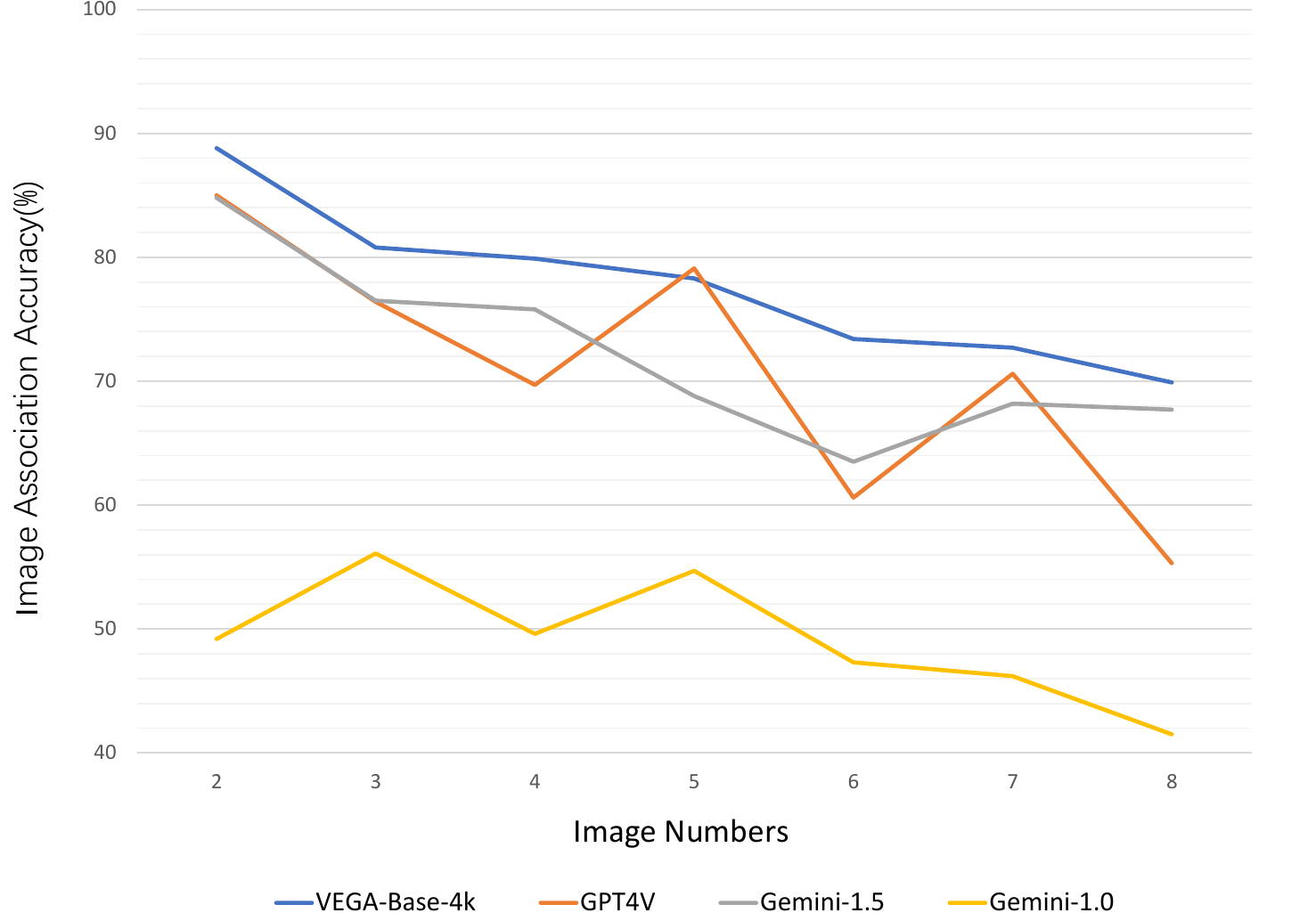}
        \caption{Variation of four models' Image Association Accuracy (\%) with increasing image number.}\label{fig:num_acc}
    \end{minipage}
\end{figure}

\begin{table*}[ht]
\centering
\small
\setlength{\tabcolsep}{4pt}        
\caption{Evaluation results for various MLLMs. Models that are not fine-tuned and have the best performances are indicated with underlines.
    }
\begin{tabular}{lccccccccccccc}
\toprule
\multirow{2}{*}{\textbf{Model}} 
& \multirow{2}{*}{\textbf{Fine-tuned}}
& \multicolumn{3}{c}{\textbf{IITC 4k}} &   
& \multicolumn{3}{c}{\textbf{IITC 8k}} &  & \multicolumn{2}{c}{\textbf{ITA}} \\
\cmidrule(r){3-5}
\cmidrule(r){7-9}
\cmidrule(r){11-12}
& & Rouge-L  & BLEU    & Acc   &  
& Rouge-L  & BLEU    & Acc   &
& 3 pic    & 5 pic      \\
\midrule
InternVL-1.5 \cite{chen2023internvl}   & w/o         
& 0.378   &   0.128 &   0.364 &  
&  0.374  &  0.126  &  0.292  &
& 0.238   &  0.014          \\
Qwen-VL-Chat\cite{Qwen-VL}   & w/o         
& 0.323    & 0.093    & 0.002   &  
& 0.250    & 0.065    & 0.001   &
& 0.043      &    0.000     \\
Gemini-1.0-pro \cite{geminiteam2024gemini}     & w/o         
& \uline{0.415}    & \uline{0.138}    & 0.649   &  
& \uline{0.404}    &\uline{0.132}    & 0.501   &
& 0.570    & 0.092    \\
Qwen-MAX \cite{QwenVLMax2024}      & w/o 
&    0.356    &    0.107    &   0.684   & 
&     0.343   &    0.107    &   0.365   &
& 0.231     &    0.017      \\
GPT4v \cite{openai2024gpt4}    & w/o  
& 0.342    & 0.091   & \uline{0.805}   &  
& 0.322    & 0.085   & \uline{0.752}   &
& 0.912    & 0.538      \\
Gemini-1.5-pro \cite{geminiteam2024gemini} & w/o    
& 0.354    & 0.087    & 0.753   &  
& 0.363    & 0.090    & 0.737   &
& \uline{0.956}    & \uline{0.891}       \\
\midrule
VEGA-Base-4k    & w
& \textbf{0.508} & \textbf{0.252}  &  \textbf{0.858} & 
& \textbf{0.488} & \textbf{0.228} & \textbf{0.789}  &
& \textbf{0.998} & \textbf{0.992}        \\
VEGA-Base-8k    & w
& 0.492    & 0.241    & 0.845   &  
& 0.473    & 0.223    & 0.775   &
& 0.994    & 0.936        \\
\midrule
Random Guess & - 
& - & - & 0.227 &
& - & - & 0.227 &
&0.167 & 0.008 \\
\bottomrule
\end{tabular}

    \label{tab:main_result}
\end{table*}

\subsection{Dataset Statistics}
\label{sec: data_statistics}

The VEGA dataset is an extensive repository featuring 50k scientific literature entries, over 200k question-and-answer pairs, and a rich trove of 400k images. It includes the IITC subset, which is segmented into two categories based on token length: one supports up to 4,000 tokens, while the other extends to 8,000 tokens. Here, images are equated to 256 tokens each. Both categories offer roughly 200k training instances and approximately 700 meticulously curated, high-caliber test samples.
The ITA task is categorized into six divisions, with two dedicated to image quantity and three to text length. Table. \ref{table:composition_vega} presents the train and test data statistics for the VEGA dataset, while Fig. \ref{fig:tokennum} details the distribution of image numbers and token counts within the IITC subset, providing insights into the visual and textual context lengths.

\subsection{Baseline Model}
\label{sec: baseline_model}
To further validate our dataset, we fine-tune the Qwen-VL-Chat~\cite{Qwen-VL} model at two distinct maximum token lengths, 4k and 8k, training a dedicated model for each configuration, denoted as VEGA-Base-4k and VEGA-Base-8k. Given the IITC task's complexity, we adopt a multi-task, multi-scale training regimen. We simultaneously train on the IITC and ITA tasks, categorizing them into varying difficulty levels based on image count and token length, as detailed in Fig. \ref{fig:tokennum}. This method aims to simplify the learning curve, bolster the model's grasp of image-text correlations, and amplify its capacity to distill essential information. Our training utilizes roughly 340K data points. For a detailed data breakdown, please see our supplementary materials.

In line with the Qwen-VL-Chat fine-tuning protocol, we freeze the visual encoder while honing the language model and adapter module. We implement a cosine learning rate schedule starting at 1 × e-5 and leverage DeepSpeed Zero-3 for distributed training.
The VEGA-Base-4k utilizes a batch size of 8, taking up 31 A100 GPU days, whereas the VEGA-Base-8k, also running with a batch size of 8, totaled 48 A100 GPU days.

%% file: sec/4_experiments.tex
\begin{table*}[h]
\centering
\small
\setlength{\tabcolsep}{4pt}          
\caption{Analysis on the effectiveness of multi-task and multi-scale training strategies. Multi-task denotes jointly training on the IITC and ITA tasks, otherwise training solely on the IITC task. Multi-scale implies employing three textual scales during the ITA task training, otherwise using only the Extent Context as a single scale.}
\begin{tabular}{cccccccccccccc}
\toprule
\multirow{2}{*}{\textbf{Multi-Task}} 
& \multirow{2}{*}{\textbf{Multi-Scale}}
& \multicolumn{3}{c}{\textbf{IITC 4k}} &   
& \multicolumn{3}{c}{\textbf{IITC 8k}} &  & \multicolumn{2}{c}{\textbf{ITA}} \\
\cmidrule(r){3-5}
\cmidrule(r){7-9}
\cmidrule(r){11-12}
& & Rouge-L  & BLEU    & Acc   &  
& Rouge-L  & BLEU    & Acc   &
& 3 pic    & 5 pic      \\
\midrule
w/o  & -       
& 0.501  & 0.247  & 0.838   &  
& 0.478 &  0.220  &  0.766  &
& -  &  -             \\
w    & w/o       
&  0.502    &   0.247     &  0.833       &  
& 0.485  & 0.228  & 0.779  &
& 0.155 & 0.004          \\
w    & w       
& \textbf{0.508} & \textbf{0.252}  &  \textbf{0.858} &  
& \textbf{0.488} & \textbf{0.228} & \textbf{0.789}&
&  \textbf{0.998}  & \textbf{0.992}              \\
\bottomrule
\end{tabular}
    \label{tab:multi_strategy}
\end{table*}

\begin{table*}[h]
\centering
\caption{Performance comparison of the base model on SciGraphQA and IITC 8k Task with different training sets. VEGA\textsuperscript{*} denotes the first context construction method for the IITC task that integrates images and their initial mentioning paragraphs from several papers, as detailed in Section \ref{subsec:Subset for IITC}.}
\begin{tabular}{lccccccccccccc}
\toprule
\multirow{2}{*}{\textbf{Base model}} 
& \multirow{2}{*}{\textbf{Training Set}}
& \multicolumn{2}{c}{\textbf{SciGraphQA}} &   
& \multicolumn{3}{c}{\textbf{IITC 8k}}  \\
\cmidrule(r){3-4}
\cmidrule(r){6-8}
&&Rouge-L  & BLEU    &  
& Rouge-L  & BLEU    & Acc     \\
\midrule
Qwen-VL-Chat   & SciGraphQA       
&  \textbf{0.538}    &    \textbf{0.290}   &           
& 0.406   & 0.153  &  0.000              \\
Qwen-VL-Chat   & VEGA      
&  0.522  & 0.266   &      
&  \textbf{0.473}  & \textbf{0.223}   & \textbf{0.775}              \\
Qwen-VL-Chat   & VEGA\textsuperscript{*}      
& 0.507   & 0.241   &      
&  0.415  &  0.160  &   0.521          \\
\bottomrule
\end{tabular}
    \label{tab:comp_scigraphqa}
\end{table*}

\section{Experiments}
\subsection{Experimental Setup}
\label{subsec:Experimental Setup}
We evaluate the performance of the current state-of-the-art MLLMs on the IITC and ITA tasks. Since most advanced open-source MLLMs do not support multi-image inputs or have not been trained on multi-image data, we only test two open-source MLLMs: 1) InternVL \cite{chen2023internvl} and 2) Qwen-VL-Chat \cite{Qwen-VL}, as well as four advanced proprietary models: 1) Gemini-1.5-pro \cite{geminiteam2024gemini}, 2) Gemini-1.0-pro \cite{geminiteam2024gemini}, 3) GPT4V \cite{openai2024gpt4}, and 4) Qwen-Max \cite{QwenVLMax2024}. Finally, we evaluate the performance of our baseline models VEGA-Base-4k and VEGA-Base-8k. During the evaluating of proprietary models, a small fraction of issues arise where the models are unable to provide answers. These instances are excluded from the scoring statistics. When evaluating Qwen-Max, the API call is limited to a maximum token input of 6000, leading to the exclusion of some test data, considering the cost of evaluating, we standardized the input images by resizing them to \(448 \times 448\) pixels.

\subsection{Main Results}
Table \ref{tab:main_result} illustrates the evaluation results of the mainstream state-of-the-art MLLMs on the VEGA dataset, highlighting the considerable challenge posed by the IITC and ITA tasks. InternVL and Qwen-VL-Chat exhibit poor performance on both tasks, which can be attributed to these open-source models' limited capability to follow instructions. Consequently, they struggle significantly with complex contextual inputs and directives inherent in these tasks.
Among the proprietary models, GPT4V achieves the highest image-relation accuracy on the IITC task. However, error analysis indicates that the primary issues include interference from similar images and instability in following instructions. Gemini-1.5-pro demonstrates exceptional accuracy in the ITA task, underscoring its robust image-text comprehension capabilities. In terms of text generation quality, Gemini-1.0-pro scores higher on Rouge-L and BLEU metrics compared to Gemini-1.5-pro and GPT4V. This discrepancy is due to two main factors: 1) Gemini-1.5-pro and GPT4V tend to summarize using their own language rather than directly incorporating descriptions from the original text, and 2) GPT4V's outputs are often longer than the Ground Truth, disadvantaging it in automated evaluations. More qualitative evaluations of these proprietary models are given in the supplementary materials.
Our model, trained using Qwen-VL-Chat 7B on the VEGA dataset, achieves state-of-the-art results across all metrics in both tasks. This demonstrates the effectiveness of our dataset in enhancing the model's capability to process interleaved image-text inputs.

\subsection{Ablation Study}
\label{subsec:Ablation Study}
\textbf{Multi-task and Multi-scale Learning.} We investigate the impact of multi-task and multi-scale training strategies on model training. As shown in Table \ref{tab:multi_strategy}, our base model is Qwen-VL-Chat, where ``Multi-Task'' represents joint training on both the IITC and ITA tasks, while ``without Multi-Task'' indicates training solely on the IITC task. ``Multi-Scale'' denotes the use of a multi-scale training strategy in the ITA task, otherwise training is only conducted on the extended context scale of the ITA task. The experimental results show that the model employing both Multi-Scale and Multi-Task strategies improves the accuracy of image-text association by 2.3\% compared to the model without these strategies, confirming the effectiveness of the multi-scale and multi-task training strategies. We also find that without multi-scale training, the model shows almost no performance improvement on the ITA task, with accuracy comparable to random guessing. 
Observing the model's failure case, it organizes the pictures and text chaotically. 
Such failure originates from two primary causes: First, the substantial volume of both text and images adds complexity to the task, challenging the model's ability to grasp the connections between textual and visual contexts. This complexity hinders the model's compliance with the directives for producing the correct output. Second, the complexity of ITA is attributed to the necessity of navigating multiple visual-textual relationships rather than identifying a single image in the answer. By adopting a multi-scale approach that streamlines the learning process during training, we have improved the model's compliance with instructions, leading to a significant enhancement in its performance on the ITA task. Additionally, we evaluate the impact of different lengths of visual context on the model performance in Fig.~\ref{fig:num_acc}.

\textbf{Comparison of SciGraphQA and VEGA.} We trained Qwen-VL-Chat using the SciGraphQA~\cite{li2023scigraphqa} and VEGA datasets and tested the models' performance on these two datasets. As shown in Table \ref{tab:comp_scigraphqa}, the model trained with the VEGA dataset performed better on both tasks, while the model trained with SciGraphQA struggled with the IITC task. These results suggest that training with the VEGA dataset not only enhanced the model's ability to handle long interleaved image-text inputs but also maintained its capability to process traditional VQA  input patterns.

\textbf{Comparison of Context Construction Methods.} We assessed the impact of two context construction techniques on model performance for the IITC task. VEGA\textsuperscript{*} represents the first method that builds multi-modal context for the IITC task by integrating images and their initial mention paragraphs from several papers, as detailed in Section \ref{subsec:Subset for IITC}. 
As shown in Table \ref{tab:comp_scigraphqa}, the data reveals that VEGA\textsuperscript{*}  underperforms on both SciGraphQA and IITC tasks compared to our baseline model, underscoring our method's superiority. This inferior performance may stem from the fact that combining text and image context from the same paper could introduce extraneous material, inadvertently sharpening the model's proficiency in discerning relevant information from noise during training.

%% file: sec/5_conclusion_limitation.tex
\section{Conclusion}
\label{sec:conclution}

In this study, we present the VEGA dataset, tailored to boost MLLMs' Interleaved Image-Text Comprehension (IITC) in real-world applications. We employ the VEGA test set to evaluate leading MLLMs, exposing the stringent requirements of IITC tasks, particularly in multi-modal comprehension with a focus on instruction adherence. Our cutting-edge approach integrates multi-task and multi-scale learning, utilizing VEGA within the Qwen-VL-Chat framework to refine MLLMs' ability to interleave image-text comprehension and accurately generate image indexes following instructions. This approach outstrips premier proprietary models like Gemini 1.5 Pro and GPT4V, proving VEGA's potential to substantially improve IITC performance. Future enhancements to the dataset will involve embedding subtler instructions, increased image-text interactivity in answers, a wider array of data sources, and more cutting-edge evaluation methods for the IITC task.

%% file: sec/6_appendix.tex



\appendix
\newpage
\onecolumn

\section{Training Data}
\begin{figure}[H]
  \centering
  \includegraphics[width=0.8\linewidth]{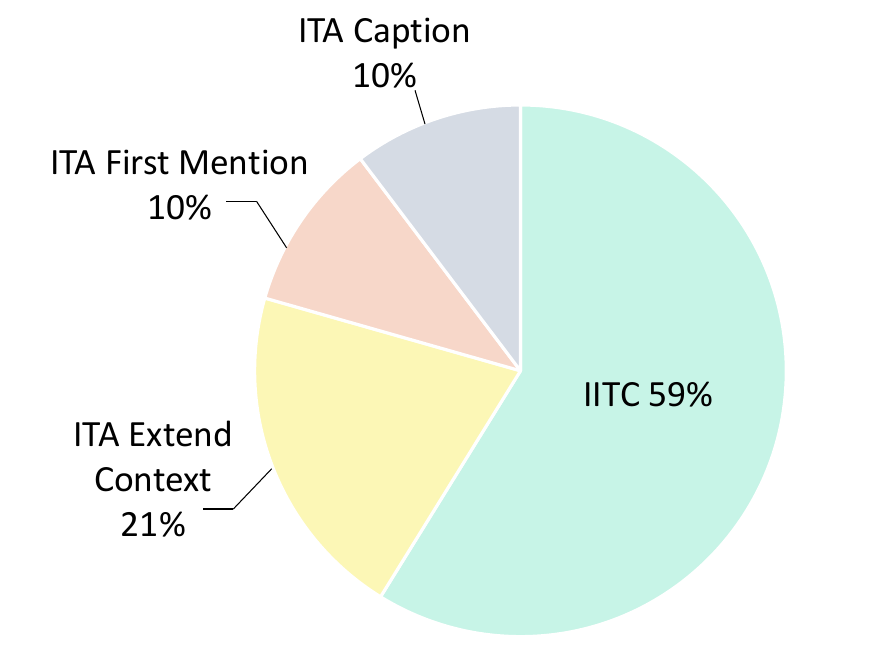}
  \caption{The composition of data for model training.}\label{fig:train_data_compo}
\end{figure}

%
\textbf{Data Composition} During the training of the VEGA-Base model, we utilized the entire IITC subset of the VEGA dataset. As for the ITA subset, we incorporated all of the Expanded Context data, as well as half of the data from both the Captions and First Mentions. The training data distribution is visualized in Figure \ref{fig:train_data_compo}.

\section{Question Filtering}
\label{sec:appendix_question_filter}
In the construction of the VEGA dataset, we filtered out the questions 
\begin{verbatim}
What are the implications of the results shown in the graph?
What are some of the potential applications of the graph?
What are some of the implications of this graph?
Are there any other interesting aspects of the graph that you would like to highlight?
What is the significance of the results shown in the graph?
What is the main goal of the experiment illustrated in the graph?
What are the key takeaways from this graph?
What are some of the key observations that can be made from the graph?
What are the key features of the mutation plot?
What are the implications of the findings in the graph?
What are the limitations of this graph?
What are the implications of the results of this graph?
Are there any limitations to the conclusions that can be drawn from the graph?
What are some of the key takeaways from this graph?
What are some of the limitations of the graph?
What are the implications of the results of the graph?
What are some of the implications of the results in the graph?
What is the significance of the different colors in the graph?
What are the limitations of the graph?
What are the implications of the results shown in the figure?
What is the main focus of the graph?
What is the significance of the 3D plot?
What is the main message of the graph?
What does the y-axis represent?
What are the key takeaways from the graph?
What can be inferred from the graph?
What is the purpose of this graph?
How does the graph support the claims made in the paper?
What are the key features of the graph?
What is the main idea of the graph?
What are the key takeaways from the figure?
What is the significance of the graph as a whole?
What is the purpose of the graph?
Are there any other interesting aspects of the graph that you would like to point out?
What is the main purpose of the graph?
What is the main goal of the graph?
What is the significance of the markers in the graph?
What are the implications of this graph for future research?
What does the graph show about the performance of the proposed method?
What is the main objective of this graph?
What is the main idea of the figure?
What is the significance of the graph in the context of the paper?
What does the x-axis represent?
What is the difference between the two figures in the graph?
What does the x-axis and y-axis of the graph represent?
What are the two main lines in the graph?
What are the implications of the results in this graph?
What does the graph show?
What is the significance of this graph?
What is the main takeaway from the graph?
What are the implications of the findings from this graph?
What are the main takeaways from the graph?
What is the purpose of the two figures in the graph?
What are the key findings of the graph?
What are the two main axes of the graph?
What do the colors in the graph represent?
What are some of the implications of the findings presented in the graph?
\end{verbatim}

\section{Case Study}
\begin{figure}[H]
  \centering
  \includegraphics[width=1.0\linewidth]{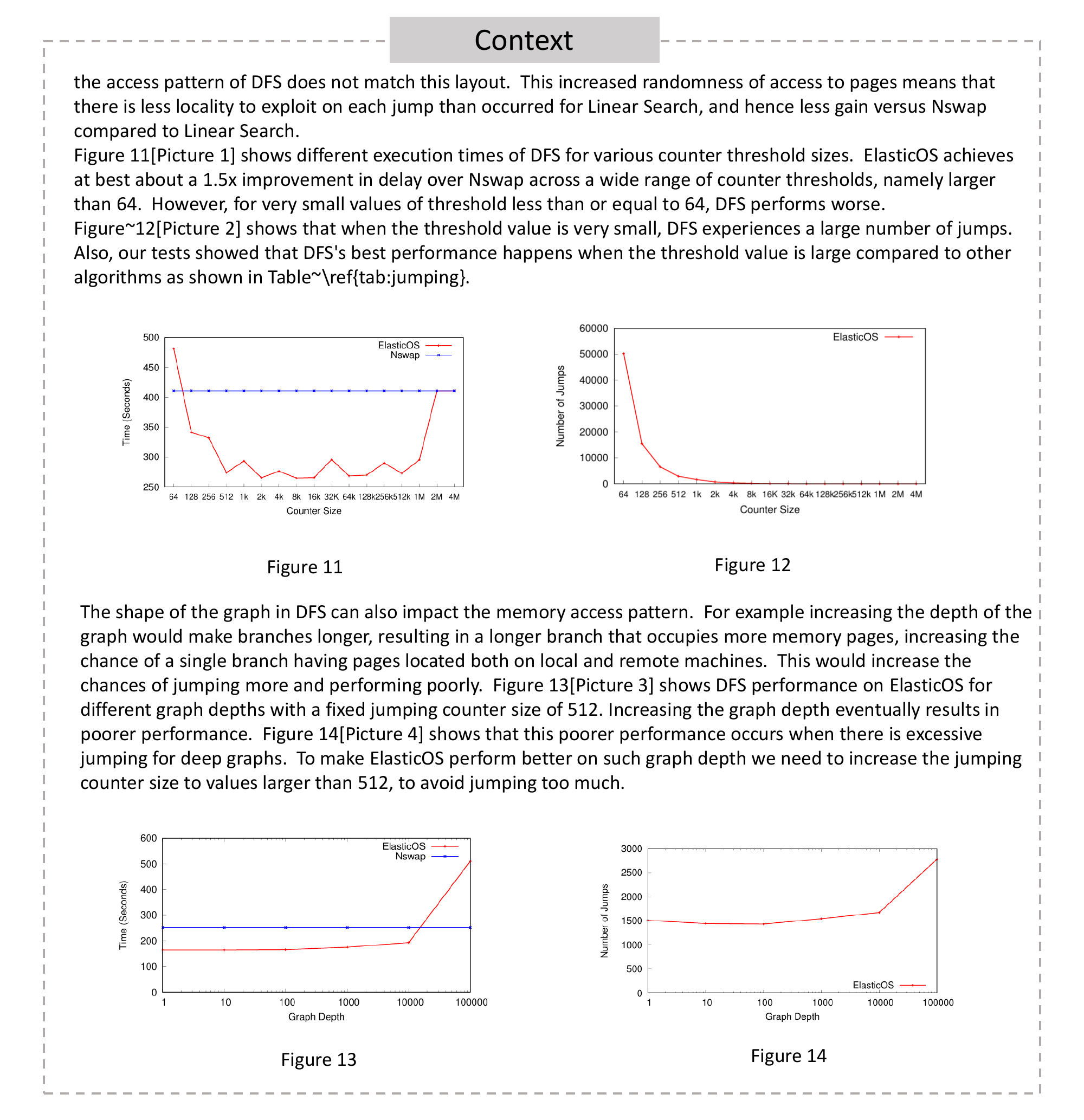}
  \caption{The context for Case 1}\label{fig/appendix3}
\end{figure}

\begin{figure}[H]
  \centering
  \resizebox{\textwidth}{!}{\includegraphics{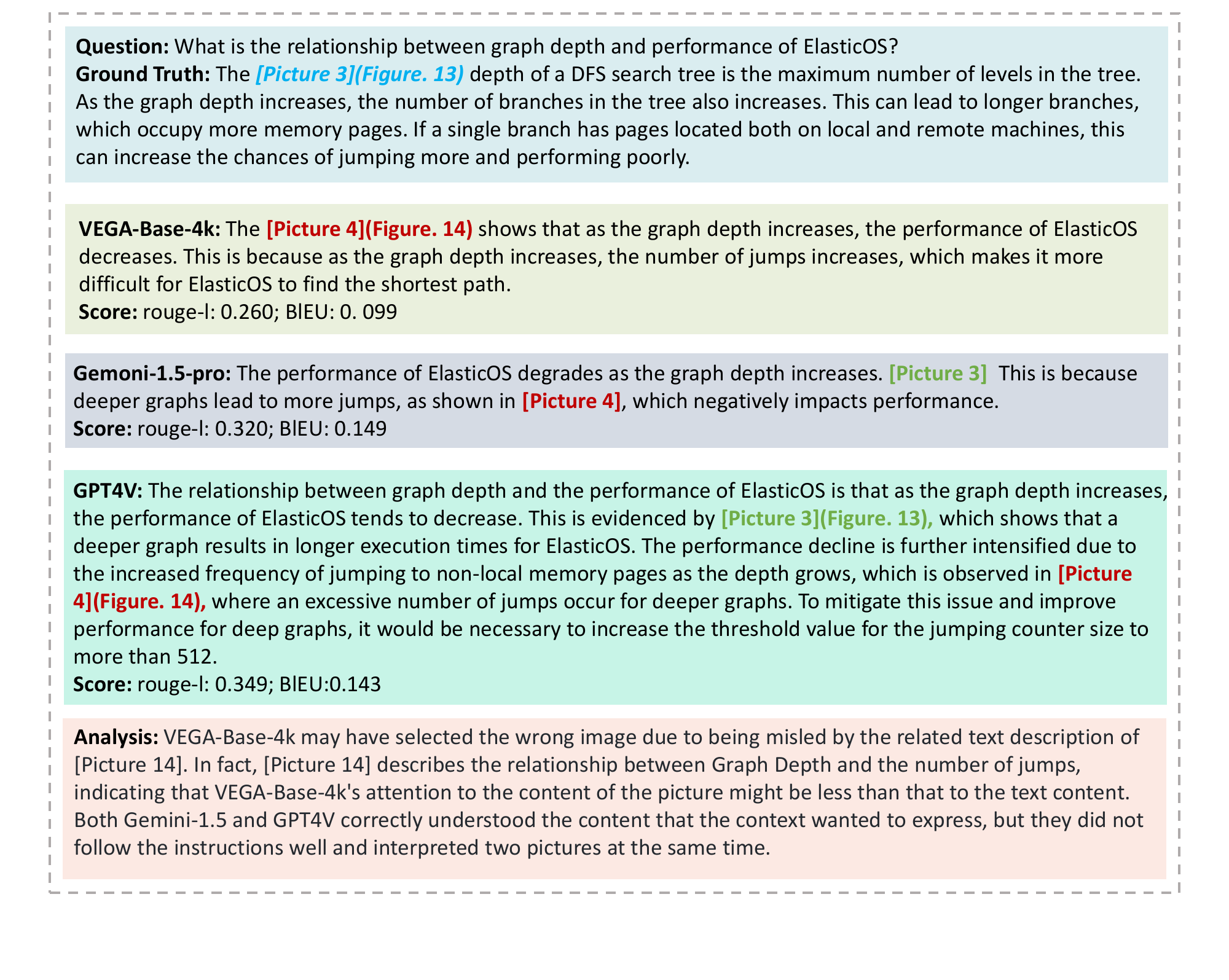}}
  \caption{Models' answer and analysis for Case 1. The pictures referred in the Ground Truth are marked in \textcolor{blue}{blue}, the incorrect images referred to in the model's answer are marked in \textcolor{red}{red}, and the correct images are marked in \textcolor{green}{green}.}
  \label{fig/appendix3ans}
\end{figure}

\begin{figure}[H]
  \centering
  \includegraphics[height=\textheight]{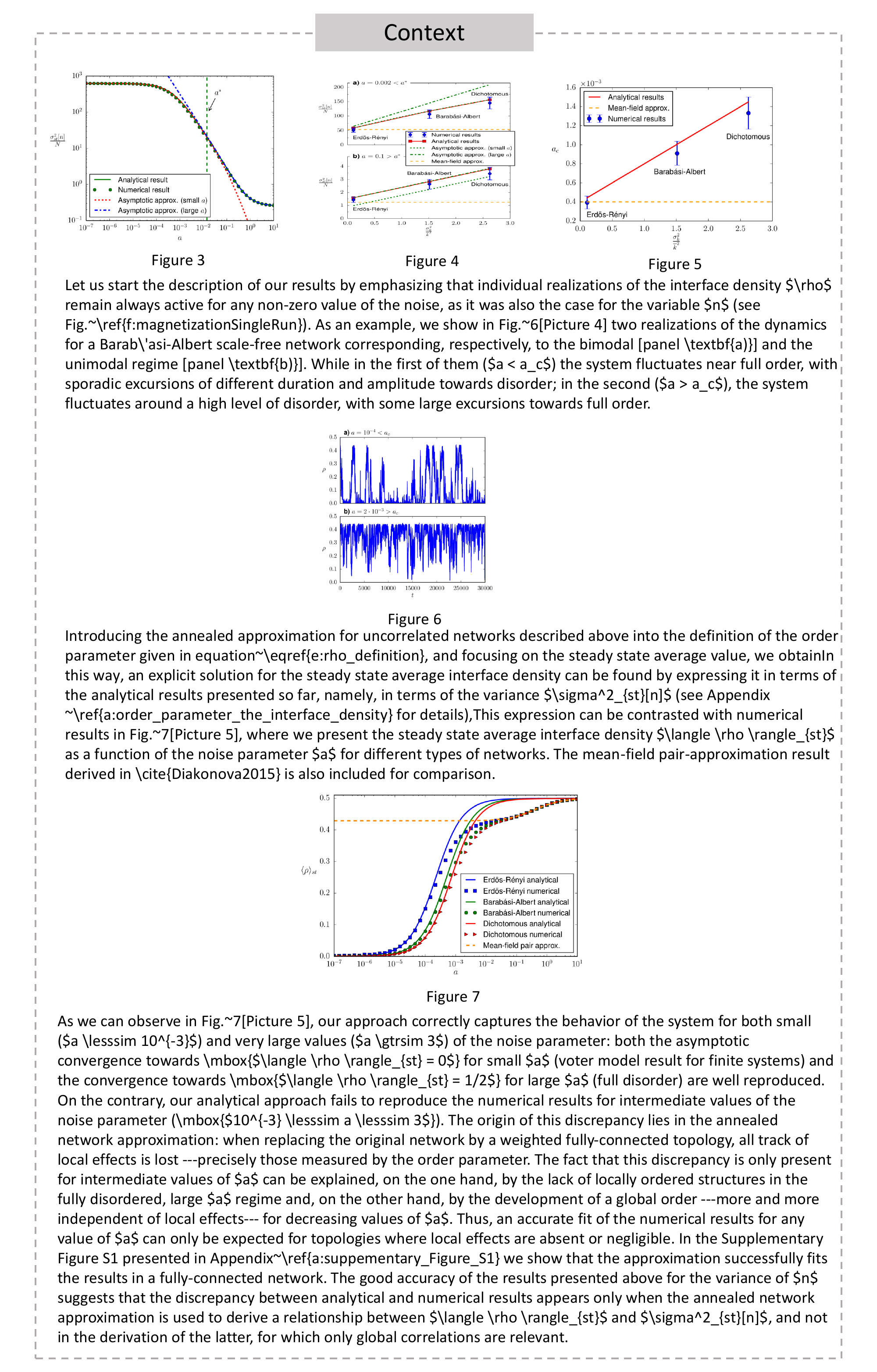}
  \caption{The context for Case 2}\label{fig/appendix4}
\end{figure}

\begin{figure}[H]
  \centering
  \includegraphics[width=1.0\linewidth]{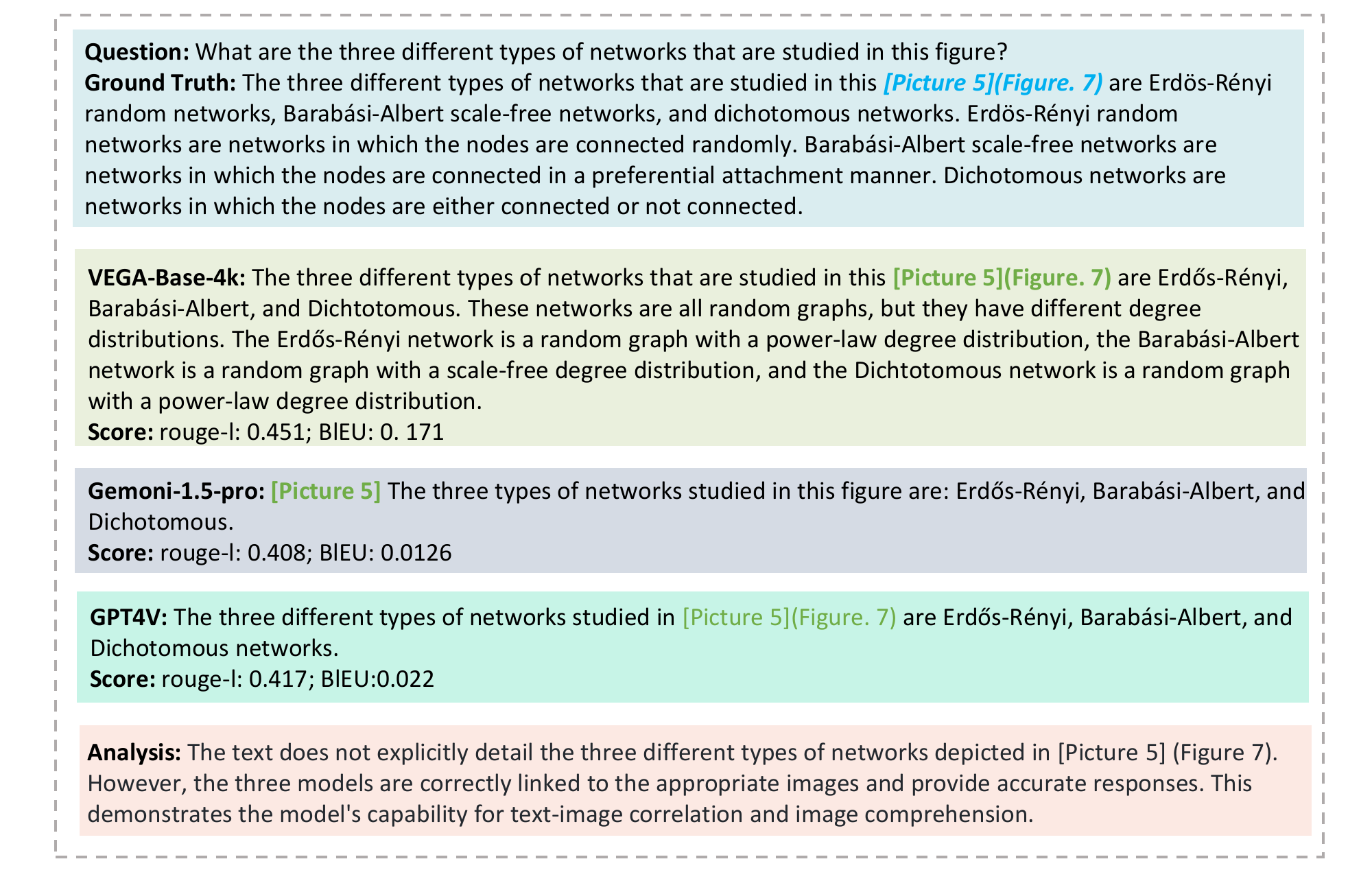}
  \caption{Models' answer and analysis for Case 2. The pictures referred in the Ground Truth are marked in \textcolor{blue}{blue}, the incorrect images referred to in the model's answer are marked in \textcolor{red}{red}, and the correct images are marked in \textcolor{green}{green}.}\label{fig/appendix4ans}
\end{figure}

\begin{figure}[H]
  \centering
  \includegraphics[height=\textheight]{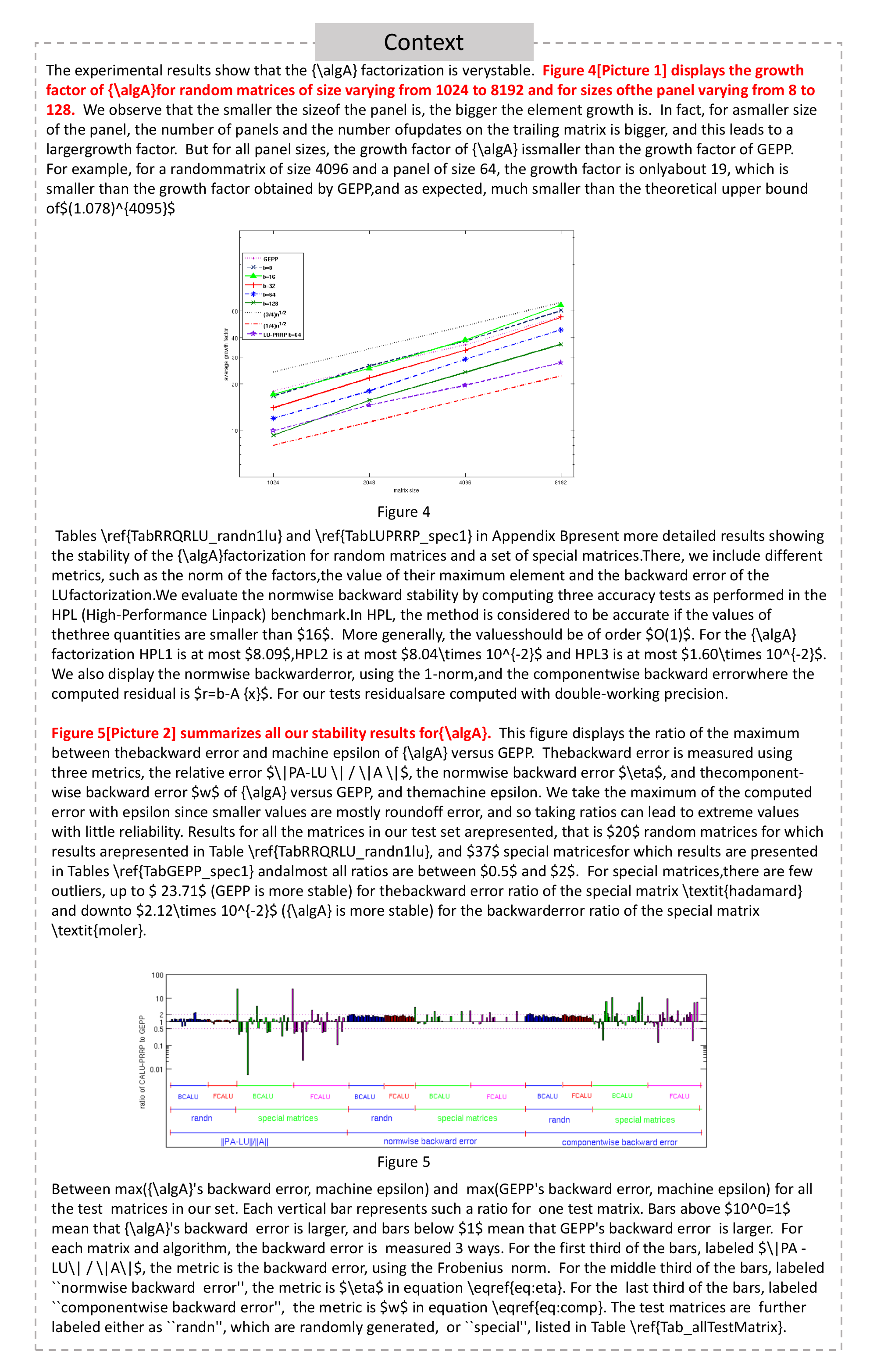}
  \caption{The context for Case 3}\label{fig/appendix1}
\end{figure}

\begin{figure}[H]
  \centering
  \includegraphics[width=1.0\linewidth]{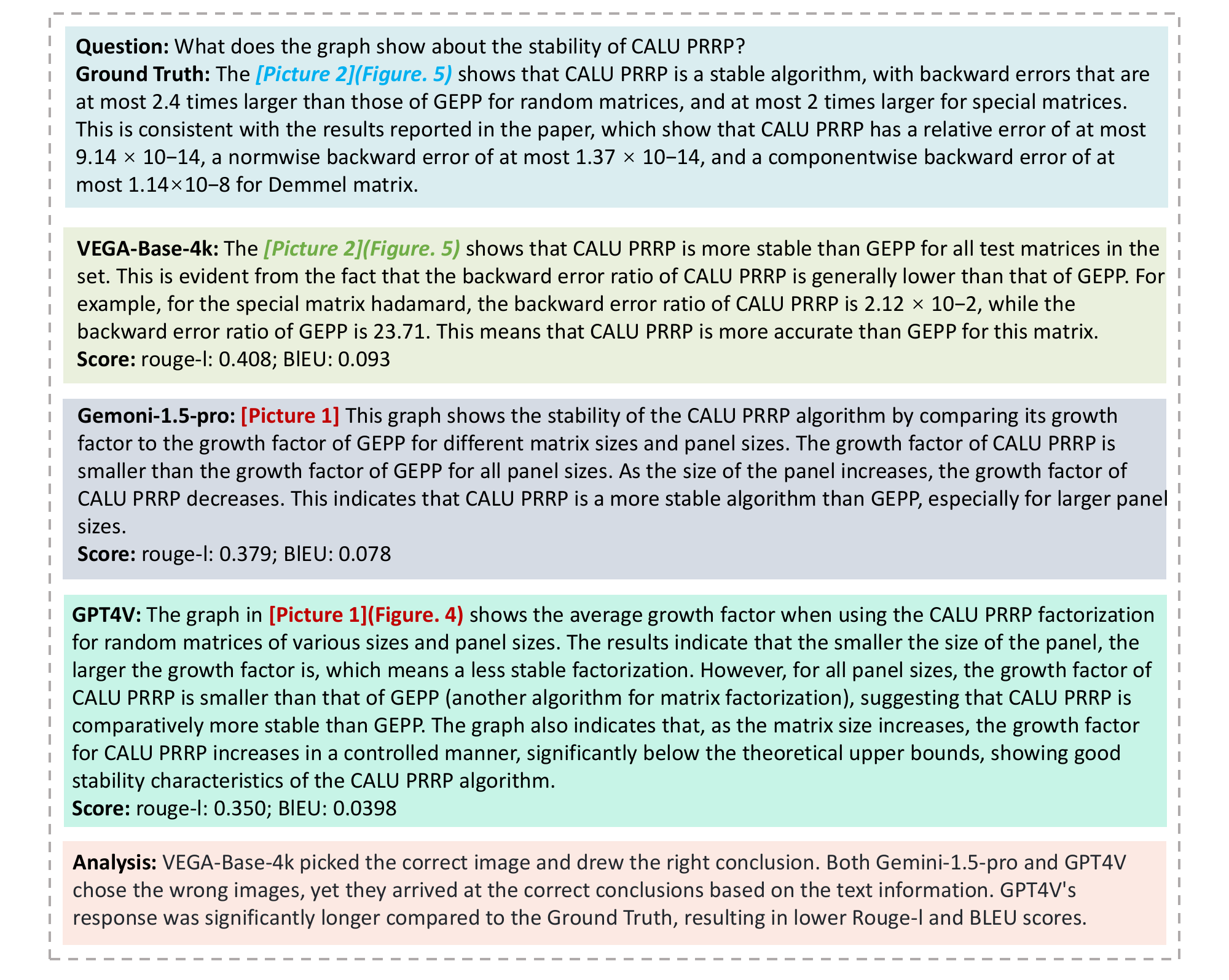}
  \caption{Models' answer and analysis for Case 3. The pictures referred in the Ground Truth are marked in \textcolor{blue}{blue}, the incorrect images referred to in the model's answer are marked in \textcolor{red}{red}, and the correct images are marked in \textcolor{green}{green}.}\label{fig/appendix1a}
\end{figure}

\begin{figure}[H]
  \centering
  \includegraphics[height=\textheight]{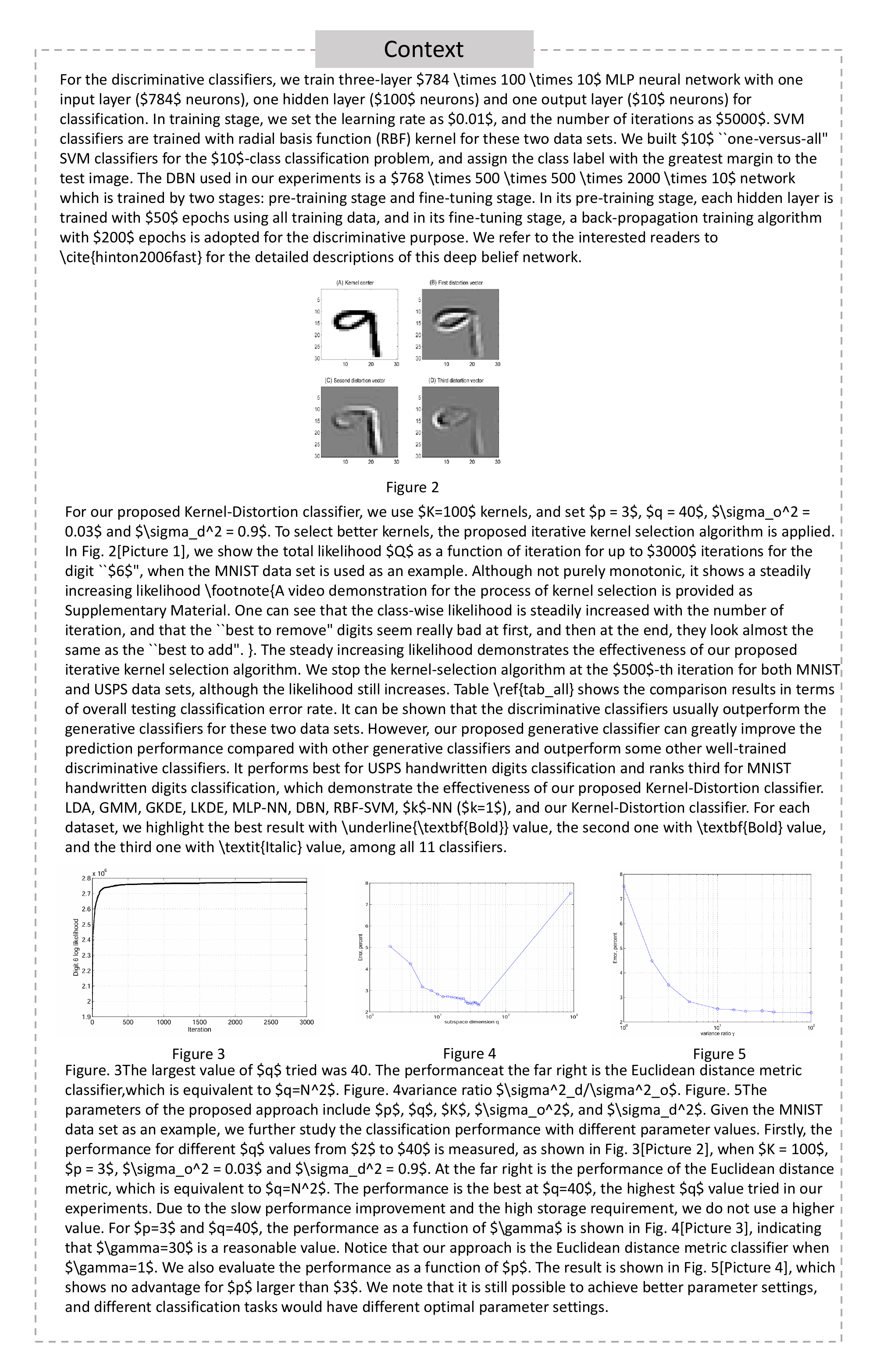}
  \caption{The context for Case 4}\label{fig/appendix2}
\end{figure}

\begin{figure}[H]
  \centering
  \includegraphics[width=1.0\linewidth]{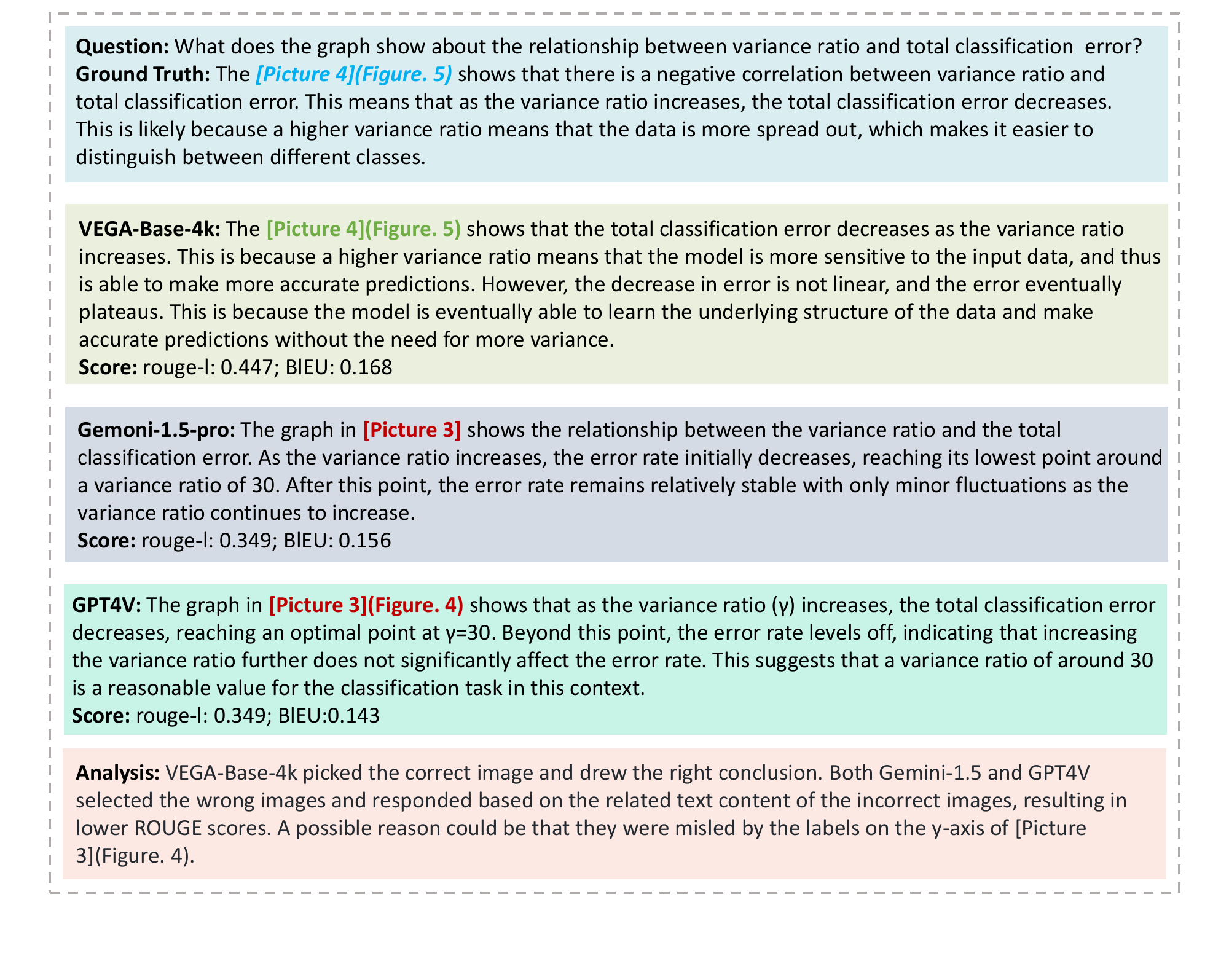}
  \caption{Models' answer and analysis for Case 4. The pictures referred in the Ground Truth are marked in \textcolor{blue}{blue}, the incorrect images referred to in the model's answer are marked in \textcolor{red}{red}, and the correct images are marked in \textcolor{green}{green}.}\label{fig/appendix2a}
\end{figure}

This section presents the performance of VEGA-Base-4k, Gemini-1.5-pro, and GPT4V on the IITC 4k task. Fig. \ref{fig/appendix3} and  \ref{fig/appendix3ans} illustrates cases where all three models failed, highlighting the challenging and complex nature of the IITC task. The models were misled by deceptive textual content, resulting in incorrect answers. Fig. \ref{fig/appendix4} and \ref{fig/appendix4ans} shows instances where all three models provided the correct response. Fig. \ref{fig/appendix1}, \ref{fig/appendix1a}, \ref{fig/appendix2} and \ref{fig/appendix2a} depict scenarios where VEGA-Base-4K answered correctly, while Gemini-1.5-pro and GPT4V did not. VEGA-Base-4k successfully navigated through distractions from similar images and irrelevant text, accurately interpreting the information in the images to arrive at the correct answers, demonstrating its robust capability to process interleaved image-text inputs.

